%% file: paper.tex
\documentclass[letterpaper, 10 pt, conference]{ieeeconf}  

\IEEEoverridecommandlockouts      
\usepackage{cite}
\usepackage{amsmath,amssymb,amsfonts}
\usepackage{algorithmic}
\usepackage{graphicx}
\usepackage{textcomp}
\usepackage{xcolor}
\usepackage{makecell}
\usepackage{array}
\usepackage[caption=false,font=footnotesize]{subfig}

\begin{document}
\makeatletter
\renewcommand\@makefntext[1]{%
  \noindent\hbox{\@thefnmark}~#1%
}
\makeatother
\title{From Language to Logic: A Theoretical Architecture for VLM-Grounded Safe Navigation}

\author{%
Kristy~Sakano$^{1}$, %
Kalonji~Harrington$^{1}$, %
and Huan~Xu$^{2}$%
\thanks{$^{1}$Department of Computer Science, University of Maryland, College Park, USA. Email: kvsakano@umd.edu and kharring@umd.edu}%
\thanks{$^{2}$Department of Aerospace Engineering, University of Maryland, College Park, USA. Email: mumu@umd.edu}%
\thanks{This work has been accepted for publication in the proceedings of the International Conference on Unmanned Aircraft Systems (ICUAS), 2026.}
}

\maketitle
\thispagestyle{empty}
\pagestyle{empty}

\begin{abstract}
We propose an architecture for integrating high-level, human-provided safety rules and operator-aligned semantic preferences into autonomous robot navigation in unstructured outdoor environments. In our approach, natural-language rules are translated into Signal Temporal Logic (STL) specifications that guide planning and navigation during runtime. Persistent, environment-centric rules and terrain preferences are grounded into a 2D cost map, while temporally dynamic requirements are expressed as STL specifications to be monitored during runtime. We hypothesize the use of Vision-Language Models (VLMs) for zero-shot scene understanding, enabling mapping between human instructions, semantic features, and environmental constraints. Within this framework, we construct an illustrative navigation model that is designed to satisfy a set of STL-encoded specifications and soft operator preferences through formal satisfaction metrics embedded into environmental properties and runtime monitoring.
\end{abstract}


\input{Sections/1_Introduction}
\input{Sections/2_Background}
\input{Sections/3_Approach}
\input{Sections/4_Conclusion}

\bibliographystyle{IEEEtran}
\bibliography{ref}

\end{document}

%% file: Sections/1_Introduction.tex
\section{Introduction}

\begin{figure}[!t]  
    \centering
    \includegraphics[width=\columnwidth]{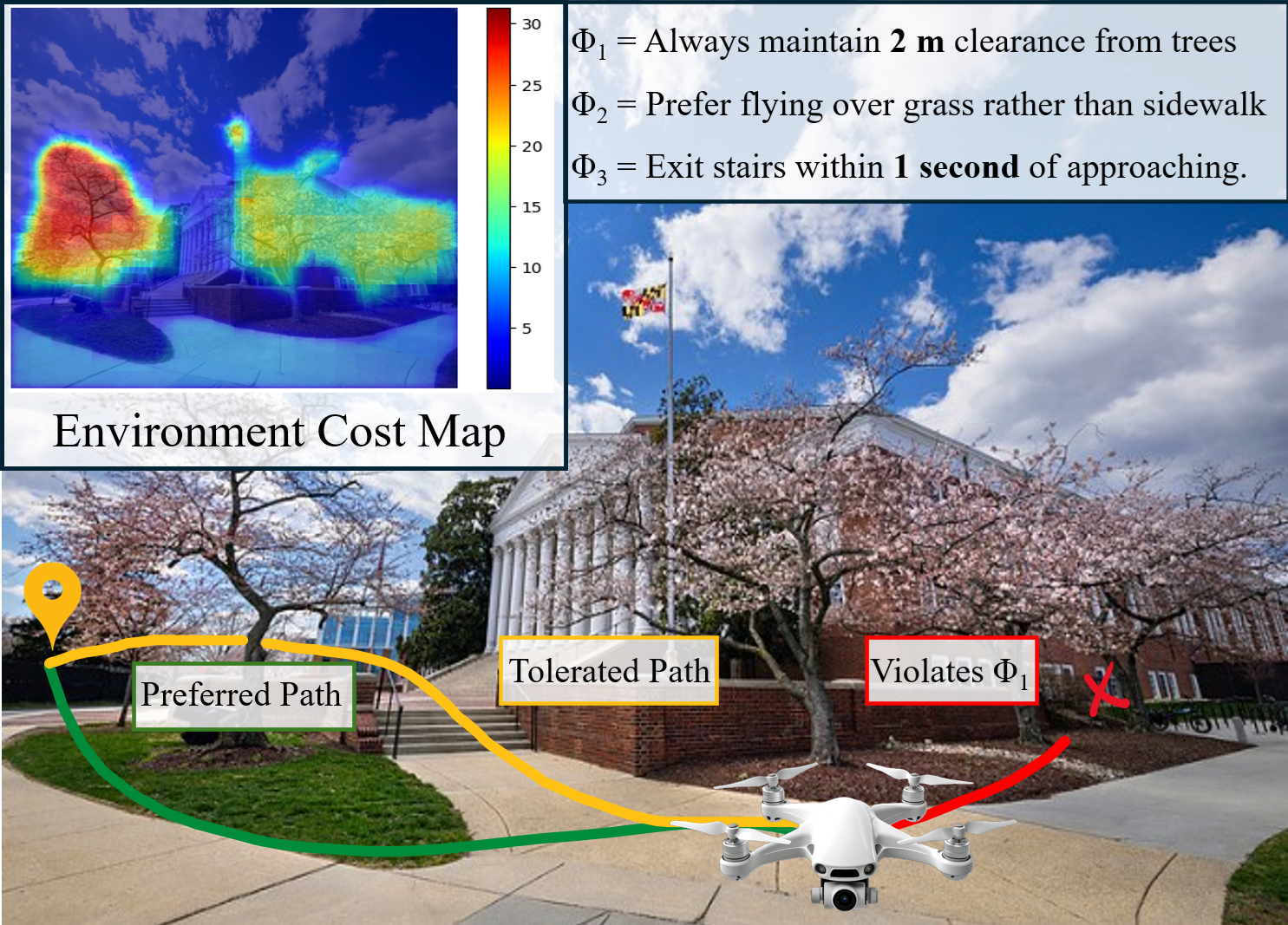}  
        \caption{Autonomous robot navigation under the proposed theoretical architecture. Operator-defined safety rules are translated into Signal Temporal Logic (STL) specifications $\phi_1, \phi_2, \phi_3$ with time-invariant preferences embedded into the environment cost map and time-varying rules integrated into the planning component and monitored at runtime.}
    \label{fig:introduction}
\end{figure}

Uncrewed Aerial Vehicles (UAVs) and other autonomous mobile robots are increasingly deployed in environments that may be novel or only partially known at deployment time \cite{ruzUAVTrajectoryPlanning2009, saleemsultanAutonomousSystemsUnstructured2024, wangSurveyPathPlanning2024}. Path planning is central to these deployments: given a representation of the environment, the planner must construct a path from the robot's current location to a goal while avoiding obstacles. For scenarios requiring high safety considerations, such as platforms operating in close proximity to humans, there is an additional requirement that the trajectory satisfy strict safety constraints and align with operating rules \cite{guiochetSafetycriticalAdvancedRobots2017}. In Figure \ref{fig:introduction}, these rules may be time-invariant, like specification $\phi_1$ \textit{``Always maintain 2 meter clearance from trees,"} or time-varying, like $\phi_3$ \textit{``Exit grass within 5 seconds upon entrance."} These rules motivate an evaluation that tracks rule satisfaction, violation, and margin along a trajectory, rather than only reporting aggregate success rates.

Ensuring safe navigation in real-world environments requires perceiving not only the geometry of surrounding obstacles, but also their semantic meaning \cite{jiaoRealTime2025}. For instance, traversing grass, water, or pavement may demand distinct behaviors depending on the associated safety risks to the robot and its surroundings. Classical navigation stacks primarily rely on dense sensing (LiDAR, depth cameras, IMUs) to reconstruct spatial maps and detect obstacles \cite{weiROSBasedNavigationObstacle2025}. However, these systems often depend on high-fidelity maps and accurate self-localization, making them brittle in unknown or unstructured environments. When using monocular, vision-based navigation, the challenge shifts to maintaining comparable safety standards with less geometric information. To navigate safely, robots must interpret the environment with semantic perception: identifying traversable regions and distinguishing objects in their environment. Classical navigation stacks are built around geometric maps and hand-engineered perception modules, so expressing rules such as \textit{``Avoid flying over red flowers, but permit flying over other flowers,"} typically require thresholds for color or task-specific segmentation models, and do not generalize well beyond predefined classes \cite{maUsingRGBImage2019, kimOpenSourceLowCostMobile2022}.

To provide this level of semantic perception on vision-based platforms, researchers have begun to explore Vision-Language Models (VLMs) that extract rich scene understanding directly from RGB observations \cite{radfordLearningTransferableVisual2021, singhFLAVAFoundationalLanguage2022}. These models combine visual input and natural language, using both to jointly identify and reason about semantically meaningful objects in an image based on high-level language instructions. Recently introduced VLMs can also provide zero-shot scene understanding, extracting useful semantic information from raw RGB images without requiring exhaustive task-specific labeling. In the context of navigation, VLMs enable open-vocabulary semantic segmentation and semantic mapping, allowing operators to describe relevant regions and objects in natural language and to ground these descriptions in the robot's visual observations. However, while VLMs expand what the robot can recognize and describe, they do not by themselves guarantee that the resulting behavior satisfies critical safety requirements.


In parallel, formal methods have been widely used to specify and verify autonomous robot behavior, from correct-by-construction motion planning with temporal logic specifications, to runtime monitoring using Signal Temporal Logic (STL) \cite{wongpiromsarnFormalMethodsAutonomous2023}. These approaches provide guarantees over complex behaviors and have been applied to both classical controllers and learning-based policies, where temporal logic bounds or evaluates behavior \cite{plakuMotionPlanningTemporallogic2014, liFormalMethodsApproach2019}. 

Despite these complementary strengths, most existing temporal logic-based specification and verification methods rely on hand-crafted predicates and pre-defined region labels, and therefore do not directly ground human language rules in raw image-based observations. Conversely, VLM-based navigation can express open-vocabulary semantic concepts but lack formalism to ensure adherence to safety constraints over time. This disconnect motivates combining temporal logic-based runtime assurance with modern VLMs for RGB-based navigation, enforcing safety rules over time in visually complex, novel environments.

\noindent \textbf{Main Contributions:} This work integrates formal evaluation into an RGB-based navigation architecture to enforce adherence to human-specified safety rules during both planning and runtime. We instantiate this framework with a theoretical navigation stack that integrates a Vision-Language Model (VLM) with Signal Temporal Logic (STL) specifications to provide runtime assurance for autonomous robots operating in unstructured outdoor environments.

\noindent \textbf{1. STL-based integration of human rules into navigation:} We translate operator-provided natural language rules into STL specifications and enforce them at two levels of the navigation stack. Time-invariant, environment-centric rules are grounded into a 2D cost map, while time-varying rules are used both for planning-time screening of candidate paths (offline robustness evaluation) and for execution-time monitoring of the candidate path (online monitoring). This dual-layer structure enables the planner to reason over both time-varying and time-invariant requirements within a unified framework.

\noindent \textbf{2. VLM grounding of formal specifications in novel environments:} We use a VLM to semantically interpret outdoor environments, applying open-vocabulary segmentation to identify relevant objects and regions described in human-provided rules. The VLM's semantic outputs encode both hard spatial constraints in the cost map and environmental properties that affect the robot's state. By grounding STL predicates directly in visual observations, the robot can reason about which regions are safe to enter or avoid, even in novel environments.

\noindent \textbf{3. Runtime assurance under changing safety requirements:} We outline a theoretical demonstration in an outdoor scenario with two operational modes: a \textit{normal mode} with baseline safety specifications and a degraded \textit{low-battery mode} with stricter requirements on speed limits, obstacle clearance, and time-bounded behavior. A mid-episode switch from the normal to degraded mode triggers recomputation of the environment cost map and re-planning of a new STL-compliant path. This scenario illustrates how tightening safety constraints and changing robot states can be handled systematically at runtime while still providing formal guarantees for autonomous navigation.

%% file: Sections/2_Background.tex
\section{Background}
Ensuring that robots act safely remains a fundamental challenge in robotics, particularly when rules are required to be followed and not embedded as only soft preferences. In this work, we focus on VLM-grounded navigation of an autonomous robot in novel environments.

We focus on a classical planning-based navigation architecture, thereby sidestepping some of the additional challenges associated with safety specification and verification of learning-based policies. Limitations of learning-based systems however, can be addressed by applying formal methods to policy synthesis, such as certificate-based verification methods,  or policy verification, such as reachability analysis\cite{manganaris2026formalmethodsrobotpolicy}. Our work is also distinct from Vision-and-Language Navigation, an extension of VLM to the robotics domain in which autonomous robots learn to follow natural language instructions to navigate novel environments. Our architecture does not align with the canonical definition of VLN, in which VLMs participate in action selection to navigate based on natural language goals. Nevertheless, the proposed framework can be layered on top of existing VLN or other vision-navigation systems as it is a safety and evaluation layer that is agnostic to the underlying policy.

In this section, we review relevant literature on (1) VLMs, (2) vision-based navigation in novel environments, and (3) the application of formal verification in autonomous robot navigation. Table \ref{tab:methods} shows a comparison between current implementations of VLNs, VLM-based navigation architectures, and non-VLM navigation architectures in the literature. 
\begin{table*}[h]
\vspace{1.5mm}
\centering
\begin{tabular}{c|c|c|c|c|c}
     \setlength{\tabcolsep}{2pt}
     \textbf{Method} & \textbf{VLM} & \textbf{Sensor Type} & \textbf{Plan Verification} & \textbf{Runtime Verification} & \textbf{Planner Type}\\
     \hline
     \hline
     
     LM-Nav \cite{shahLMNavRoboticNavigation2022} & Recognition & Camera & None & None & Learning-based \\[0.5em]

     VL-Nav \cite{duVLNavRealtimeVisionLanguage2025} & Detection \& Segmentation & Camera \& LiDAR & None & None & Classical \\[0.5em]
    
     Behav \cite{weerakoonBehavBehavioralRule2025} & Segmentation \& Goal & Camera \& LiDAR & None & None & Classical \\[0.5em]

     DV-VLM \cite{liDVVLNDualVerification2026} & Action Selection & Camera & VLM & None & None \\[0.5em]

    Baird and Coogan \cite{bairdRuntimeAssuranceSignal2023} & None & State signals        & STL (finite horizon) & STL (finite horizon) & Classical      \\[0.5em]
    
    Zero-Shot STL \cite{liuZeroShotTrajectoryPlanning2025}  & None & State (abstract)     & STL  (task spec)     & None                  & Learning-based \\[0.5em]
    
    STL PI Planner \cite{halderTrajectoryPlanningwithSTL2025} & None & Continuous state    & STL   (traj. cost)  & None                  & Classical      \\[0.5em]
    
    PASTEL \cite{kapoorLogicallyConstrainedRobotics2024a}   & None & State features       & STL (finite horizon) & None                  & Learning-based \\[0.5em]
    
    S-MSP \cite{yeBridgingPerceptionPlanning2025}           & None & Multiple Cameras & STL (finite horizon) & None                  & Learning-based \\ [0.5em]
     \hline

     Ours & Segmentation & Camera & STL (env. cost map) & STL (finite horizon) & Classical\\
\end{tabular}
\vspace{2mm}
\captionsetup{format=plain}
\caption{Comparison between our architecture and representative navigation approaches. Columns indicate the role of VLMs, sensing modality, whether STL or other formal verification is used at planning or run time, and planner type. }
\label{tab:methods}
\end{table*}
\subsection{Vision-Language Models (VLMs)}

Recent advances in vision-language modeling have produced VLMs that serve as strong foundation models for solving a variety of text and image tasks \cite{zhangVisionLanguageModelsVision2024, liSurveyStateArt2025}. These models either jointly learn text and image encoders to align their embeddings \cite{radfordLearningTransferableVisual2021, singhFLAVAFoundationalLanguage2022} or map image inputs to a pretrained Large Language Model (LLM) \cite{LiuLLavaVisualInstructionTuning2023, baiQwen25VLTechnicalReport2025}. Similar to LLMs, their broad pre-training dataset and multi-modal objective allows the model to incorporate rich semantic relationships critical for downstream tasks. 

One application well suited for VLMs is open-vocabulary semantic segmentation, which identifies and assigns portions of an image to arbitrary semantic classes, usually at the pixel level  \cite{ZhaoOpenVocab, fengVisionLanguageModelObjectSegmentation2025}. Using the strong semantic link between language and images found in pre-trained VLMs, recent methods have demonstrated impressive generalization to unseen categories and complex, open-world scenes.  \cite{xuGroupViTSemanticSegmentation2022, choCATSegCostAggregation2024, shiLLMFormerLargeLanguage2025}. In our setting, we use open-vocabulary segmentation to derive semantic labels and associated costs in a 2D environment map, bridging language-specified rules and the vision-based navigation stack.

\subsection{Vision-based navigation in novel environments}

The success of vision-language models in scene understanding has motivated their use in autonomous navigation \cite{radfordLearningTransferableVisual2021, zhangVisionLanguageModelsVision2024, liSurveyStateArt2025}, particularly in visually complex or previously unseen environments. Prior work has incorporated deep semantic segmentation and metric-semantic mapping to identify traversable terrain and obstacles from RGB images, enabling navigation in novel outdoor settings \cite{huangVisualLanguageMaps2023, sahuAnyTraverseOffroadTraversability2025}. In parallel, recent systems incorporate VLMs to bias classical planners, construct semantic maps, and guide exploration by highlighting semantically relevant regions, even when the goal is not directly visible \cite{duVLNavRealtimeVisionLanguage2025, songVLMSocialNavSociallyAware2025}.

For example, VLM-RRT uses VLM-derived semantic knowledge to bias sampling in a top-down map view, accelerating convergence of sampling-based planners while assuming full observability and focusing primarily on obstacle avoidance \cite{yeVLMRRTVisionLanguage2025, lavalleRapidlyExploringRandomTrees1998}. Other systems fuse VLM-based semantic embeddings directly into 3D maps to ground high-level concepts in the navigation representation, thereby supporting goal-directed navigation based on semantic landmarks \cite{huangVisualLanguageMaps2023}. These methods demonstrate the utility of VLMs for vision-based navigation in novel environments, but they typically encode safety behavior implicitly through time-invariant cost shaping or heuristics and lack explicit temporal reasoning or formal guarantees of rule compliance.





\subsection{Formal verification of autonomous navigation}

Formal methods provide a complementary perspective by specifying desired robot behaviors either during controller synthesis or at runtime \cite{wongpiromsarnFormalMethodsAutonomous2023}. Temporal logic and related verification tools enable reasoning about the evolution of robot behavior over time. In particular, STL is an expressive formalism for specifying temporal properties of continuous-time, real-valued signals \cite{malerMonitoringTemporalProperties2004a}. STL specifications combine atomic predicates, Boolean operators, and bounded temporal operators such as globally (G), until (U), and eventually (F) to express mathematically precise specifications of control systems \cite{donzeSignalTemporalLogic2013, donzeRobustSatisfactionTemporal2010}. Unlike Linear Temporal Logic (LTL), which is evaluated over discrete state sequences, STL is interpreted over dense-time systems and is therefore well suited for analyzing safety and performance in cyber-physical systems.

Prior work has employed STL for runtime assurance by formulating a finite-horizon optimization problem that minimally adjusts the nominal input while maintaining satisfaction of an STL safety specification \cite{bairdRuntimeAssuranceSignal2023}. More recent work has incorporated STL into trajectory planning with formal satisfaction guarantees. For example, Liu et. al (2025) proposed a hierarchal, data-driven framework that decomposed STL tasks into timed sub-goals and used a diffusion model to stitch together feasible trajectory segments that satisfy the specification \cite{liuZeroShotTrajectoryPlanning2025}. 

Complementary sampling-based methods integrate STL robustness directly as a cost in model predictive path integral control, solving finite-horizon optimal control problems with STL objectives \cite{halderTrajectoryPlanningwithSTL2025}. In parallel, end-to-end architectures have emerged that map raw perceptual inputs and STL specifications directly to trajectories using transformers with STL-structured conditioning \cite{kapoorLogicallyConstrainedRobotics2024a, yeBridgingPerceptionPlanning2025}. Across these methods, predicates are typically defined over structured state spaces with fixed region labels; they do not address grounding temporal logic predicates directly in open-vocabulary visual observations, nor do they explicitly separate hard safety constraints from soft operator preferences.


%% file: Sections/3_Approach.tex
\section{Approach}
In this section, we outline the methodology of our approach, structured into four components: (1) operator-provided hard constraints and soft preferences, (2) VLM-grounded navigation with hard constraints, (3) state change, and (4) Test and Evaluation (T\&E) roadmap. 

\begin{figure*}[tbp]  
\vspace{2mm}
    \centering
    \includegraphics[width=1.9\columnwidth]{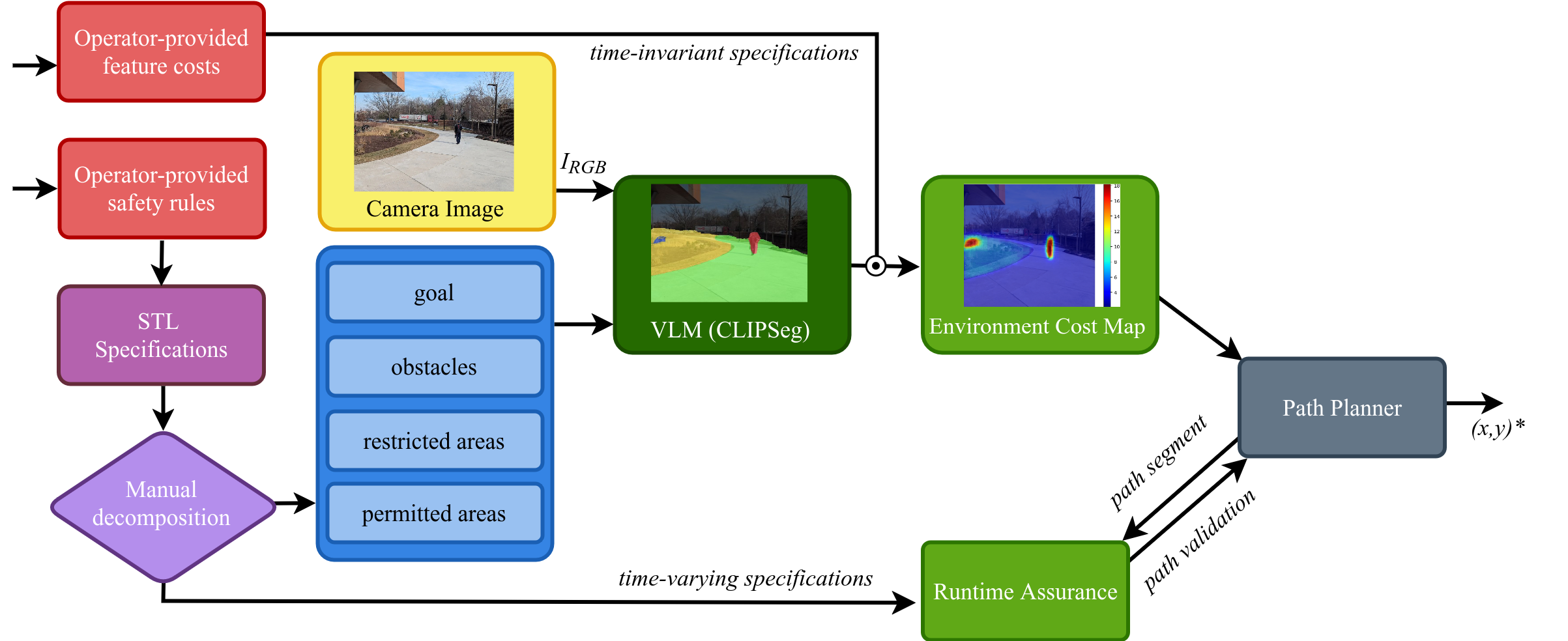}  
    \caption{Overall theoretical architecture of our VLM-grounded safe navigation stack. We obtain \textit{operator-provided feature costs} and \textit{operator-provided safety rules} from mission specifications and hand-translate the rules into \textit{STL specifications} that are manually decomposed into relevant features for the \textit{VLM} to parse in a \textit{RGB camera image}. Combined with the \textit{feature costs}, we construct an \textbf{environment cost map} for all temporally-invariant specifications. The time-varying specifications are incorporated into a \textbf{runtime assurance} layer that iteratively validates paths proposed by the \textit{path planner}.}
    \label{fig:approach}
\end{figure*}

\subsection{Operator-Provided Hard Constraints and Soft Preferences}
Our architecture assumes that operators specify both hard constraints and soft preferences prior to runtime. Hard constraints represent safety-critical requirements that must not be violated (e.g., speed limits, obstacle clearnace, time-bounded behavior) while soft preferences encode non-critical choices (e.g., preferring flight over grass rather than above sidewalks) \cite{baierPlanningPreferences2008}. Soft preferences are embedded as static costs in the environment cost map, and hard constraints are encoded both in the cost map (for time-invariant spatial rules) and in STL specifications used during planning and runtime.

\subsubsection{Hard Constraints to STL Specifications}
Human operators oversee verification and operation of autonomous robots and ensure compliance with safety standards prior to deployment. Natural language descriptions of hard constraints are translated into STL specifications; the details of this translation process are abstracted from this paper, and we assume the correctness of the translation. Related work demonstrates how natural language processing techniques can be used for transformations \cite{heDeepSTLEnglishRequirements2022, maoNL2STLTransformationLogic2024, xuLearningFailuresTranslation2024, chenNL2TLTransformingNatural2024}.

STL is chosen for its ability to capture temporal relations among continuous system signals, an essential property for autonomous robots that continuously generate rich sensory data streams \cite{malerMonitoringTemporalProperties2004a, donzeSignalTemporalLogic2013}. Furthermore, STL's quantitative semantics allow for graded assessment of specification satisfaction or violation via robustness metrics \cite{donzeRobustSatisfactionTemporal2010}. Unlike aggregate metrics such as the percentage of time spent in hazardous regions, which ignore when and how unsafe states occur, STL specifications explicitly encode ordering, timing, and duration constraints. For example, Rule 3 in Table \ref{table:rules}, the \textit{``Sidewalk-Exit Temporal Rule"}, requires the robot to exit grass within 5 seconds of entry, permitting frequent but short excursions into grass, while penalizing prolonged grass traversal.

\subsubsection{Embedding Soft Preferences}
Soft operator constraints are embedded into the environment cost map as semantic costs. For example, Rule 5 in Table \ref{table:rules},  \textit{``Preference to avoid pedestrian sidewalks"}, specifies that the operator prefers flight paths over grass rather than above sidewalks. This preference is implemented as a relative cost difference between grass and sidewalk labels and does not appear as a hard STL constraint. 

\subsection{VLM-Grounded Navigation with Hard Constraints}
Our approach embeds the STL specifications into the navigation stack in two sections:

\noindent \textbf{1. Persistent, time-invariant rules} are enforced through spatial representations in the environment cost map, 

\noindent \textbf{2. Dynamic, time-varying rules} are used to evaluate candidate plans produced by the path planner and also continuously monitored during runtime to verify that the evolving system behavior remains within acceptable boundaries (e.g., speed limits, time-bounded responses to environmental conditions), as illustrated in prior works \cite{wongpiromsarnFormalSynthesisEmbedded2011, bairdRuntimeAssuranceSignal2023}.


\subsubsection{Environment Cost Map}
We encode time-invariant spatial rules and soft operator preferences in a 2D environment cost map by combining VLM-based semantic segmentation with operator-provided semantic costs. In particular, an open-vocabulary segmentation model such as CLIPSeg \cite{luddeckeImageSegmentationUsing2022} produces, for each image pixel, a vector of probabilities over the semantic labels referenced in the operator rules (e.g., \textit{water}, \textit{tree}, \textit{grass}, or \textit{sidewalk}).

Let $\mathcal{I} = \{1,\dots,H\} \times \{1,\dots,W\}$ denote the image grid, and let $\mathcal{L} = \{\ell_1,\dots,\ell_n\}$ denote the set of all semantic labels relevant to motion planning and our rules/preferences. We define the pixel-wise semantic segmentation function
\begin{equation}
\mathrm{Seg} : \mathcal{I} \rightarrow \mathbb{R}^n,
\end{equation}
where $\mathrm{Seg}(i,j)$ is a length-$n$ vector whose $k$-th component gives the confidence that pixel $(i,j)$ belongs to semantic label $\ell_k \in \mathcal{L}$.

We can stack these per-pixel outputs into a tensor
\begin{equation}
\mathbf{S} \in \mathbb{R}^{H \times W \times n},
\end{equation}
where $S_{ijk}$ denotes the $k$-th component of $\mathrm{Seg}(i,j)$. Thus, $\mathrm{Seg}(i,j)$ and $\mathbf{S}$ represent the same semantic information at two different levels of notation: $\mathrm{Seg}(i,j)$ refers to the semantic label vector at a single pixel, while $\mathbf{S}$ denotes the full segmentation tensor over the image.

Let $\mathbf{c} \in \mathbb{R}_{\geq 0}^{n}$ denote the vector of operator-provided costs associated with the semantic classes in $\mathcal{L}$. These costs encode both hard spatial penalties and soft terrain preferences. For example, \textit{obstacle} or \textit{tree} labels may be assigned very high costs, while \textit{grass} may be assigned a lower cost than \textit{sidewalk} to reflect a soft operator-provided preference for flying over grass rather than above pedestrian walkways.

We define the 2D environment cost map $\mathbf{C} \in \mathbb{R}^{H \times W}$ by taking the inner product of the semantic label vector at each pixel with the cost vector:
\begin{equation}
C_{ij} = \sum_{k=1}^{n} S_{ijk} \, c_k,
\label{eq:costmap_entry}
\end{equation}
for all $(i,j) \in \mathcal{I}$. In compact notation, we write
\begin{equation}
\mathbf{C} = \mathbf{S} \cdot \mathbf{c},
\label{eq:costmap_tensor}
\end{equation}
where “$\cdot$” denotes contraction of the label dimension of $\mathbf{S}$ with the vector $\mathbf{c}$. Equivalently, each entry $C_{ij}$ is the expected semantic traversal cost at image location $(i,j)$ under the segmentation output.

In practice, both the motion planner and the runtime monitor operate on trajectories represented in the 2D planning frame. The planner uses the cost map $\mathbf{C}$ to generate candidate paths that avoid high-cost regions. For each candidate path, we view the sequence of waypoints and associated state variables (e.g. distance to the nearest obstacle or speed) as a discrete-time trace and evaluate its robustness with respect to previously defined STL specifications. This provides a quantitative measure of rule adherence for the planned trajectory. If a candidate path violates any hard constraint rules (negative robustness beyond acceptable limits), the affected portion of the environment cost map is re-weighted and the planner is invoked again for a path segment repair. This iterative process continues until a candidate path satisfies all hard constraint rules, at which point robot execution can begins.

As a concrete example, consider Rule 2 in Table \ref{table:rules}, \textit{Obstacle Avoidance with Buffer}. Because this rule is time-invariant, it is encoded directly in the cost map by assigning high costs to obstacle-labeled pixels and to neighboring pixels within the desired safety margin. The planner then treats these regions as expensive to traverse, biasing paths away from obstacles before any temporal monitoring is applied. 

\subsubsection{Planning-time STL screening (offline)}
We use STL specifications in a planning-time role by screening candidate trajectories before execution. Each candidate trajectory produced by the path planner is discretized into a trace of semantically relevant states, such as approaching stop signs, entering obstacle-dense regions, or transitioning from grass to path. In our initial implementation, this segmentation is configured manually based on task structure and map annotations; in future work, we aim to use VLMs to automatically identify and segment regions of interest by aligning key phrases in the STL rules (e.g., \textit{grass} or \textit{stop sign}) with image regions.

Given a candidate trace and the set of STL specifications, we evaluate robustness offline. If a segment of the trajectory is found to have negative STL robustness, indicating a violation, we invoke an incremental, repair-capable variant of our planner that selectively modifies only the offending segment of the path rather than replacing the entire path. This allows us to reuse non-violating segments of the path and reduce computational overhead while maintaining specification compliance. 

If no feasible local repair can be found under the current specification, the system escalates to repairing a n increasingly greater subset of the candidate path, up to a full replan from the robot's current state using the existing STL constraints.

As an example, consider Rule 3, \textit{Grass-Exit Temporal Rule}, in Table \ref{table:rules}. Any candidate trajectory that remains in a grass-labeled region for more than 5 seconds after entering grass violates this specification and incurs a large penalty in the planning cost. When violation occurs, the corresponding segment is marked for repair, and the planner attempts to locally replace it with an alternative segment that meets specification. 

\subsubsection{Execution-time STL monitoring (online)}
In addition to planning-time screening, we provide execution-time monitoring for online runtime assurance. During execution, the robot's state over time is recorded as a discrete-time trace. The same STL specifications are evaluated online, producing robustness scores that indicate the degree to which each rule is currently satisfied as the robot navigates the environment.

In simulation, the executed trajectory should exactly follow a rule-satisfying plan. However, real robots in complex environments experience disturbances, perception errors, or modeling inaccuracies may cause deviations that lead to specification violations. When a violation is detected, the system flags the event and triggers replanning under the current STL constraints, using the updated environment cost map.

\subsubsection{Path Planning}
Path planning is performed using a classical search-based algorithm that optimizes a cost combining environment costs and penalties for violating STL rules. As the robot navigates through the environment, new visual information continuously refines the cost map, requiring the planner to adapt to an uncertain and dynamic environment that may include unexpected obstacles or terrain changes. 

At each planning cycle, the robot updates its environment cost map  $\mathcal{C}$ from the latest RGB observation using the VLM-based segmentation model described above. A sampling-based planner then proposes candidate trajectories in the 2D planning frame, optimizing a cost that combines $\mathcal{C}$ with penalities for STL rule violations. 

We envision using a modified version of RRT known as RRT-X, an asymptotically optimal sampling-based motion planning algorithm for real-time navigation in dynamic environments with unpredictable obstacles \cite{otteRRTXAsymptoticallyOptimal2016, otteRRTXRealTimeMotion}. Assuming that the cost map and feasible regions frequently change with new VLM detections, mode switches, and other updates, a continuous-space, kinodynamic planner can repair a single tree faster than replanning from scratch. The motion planner maintains a continuous-space tree that is incrementally repaired as specifications or semantic labels change over time.

The integration of spatial and temporal reasoning within the same planning loop ensures that generated trajectories are both \textit{valid} and \textit{formally safe}.

\subsection{State-Change: Normal versus Low Battery Mode}
\begin{figure}[!t]
\vspace{2mm}
  \centering
  \subfloat{%
    \includegraphics[width=0.9\columnwidth]{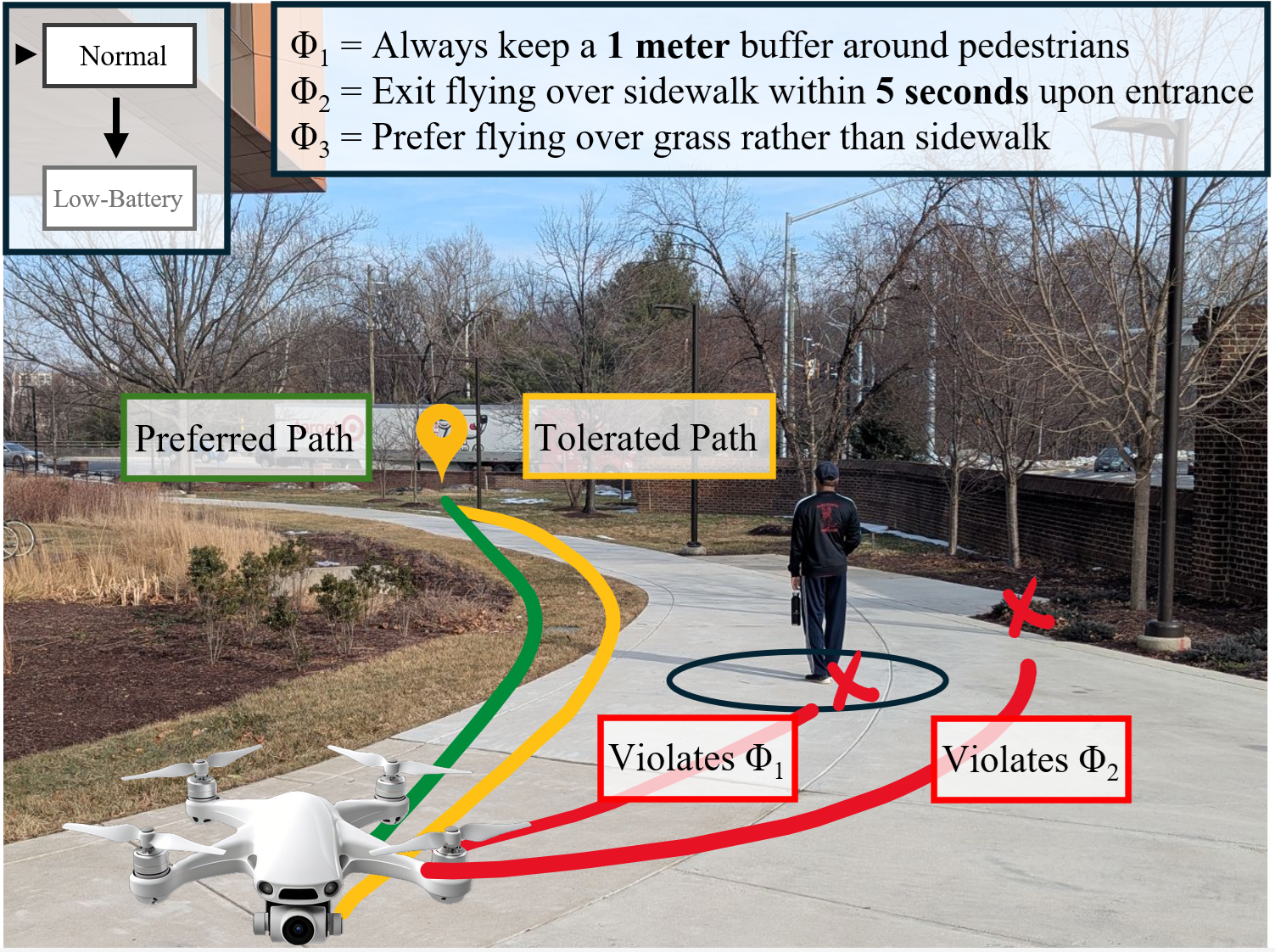}%
    \label{fig:top}
  }\\[0.5em]
  \subfloat{%
    \includegraphics[width=0.9\columnwidth]{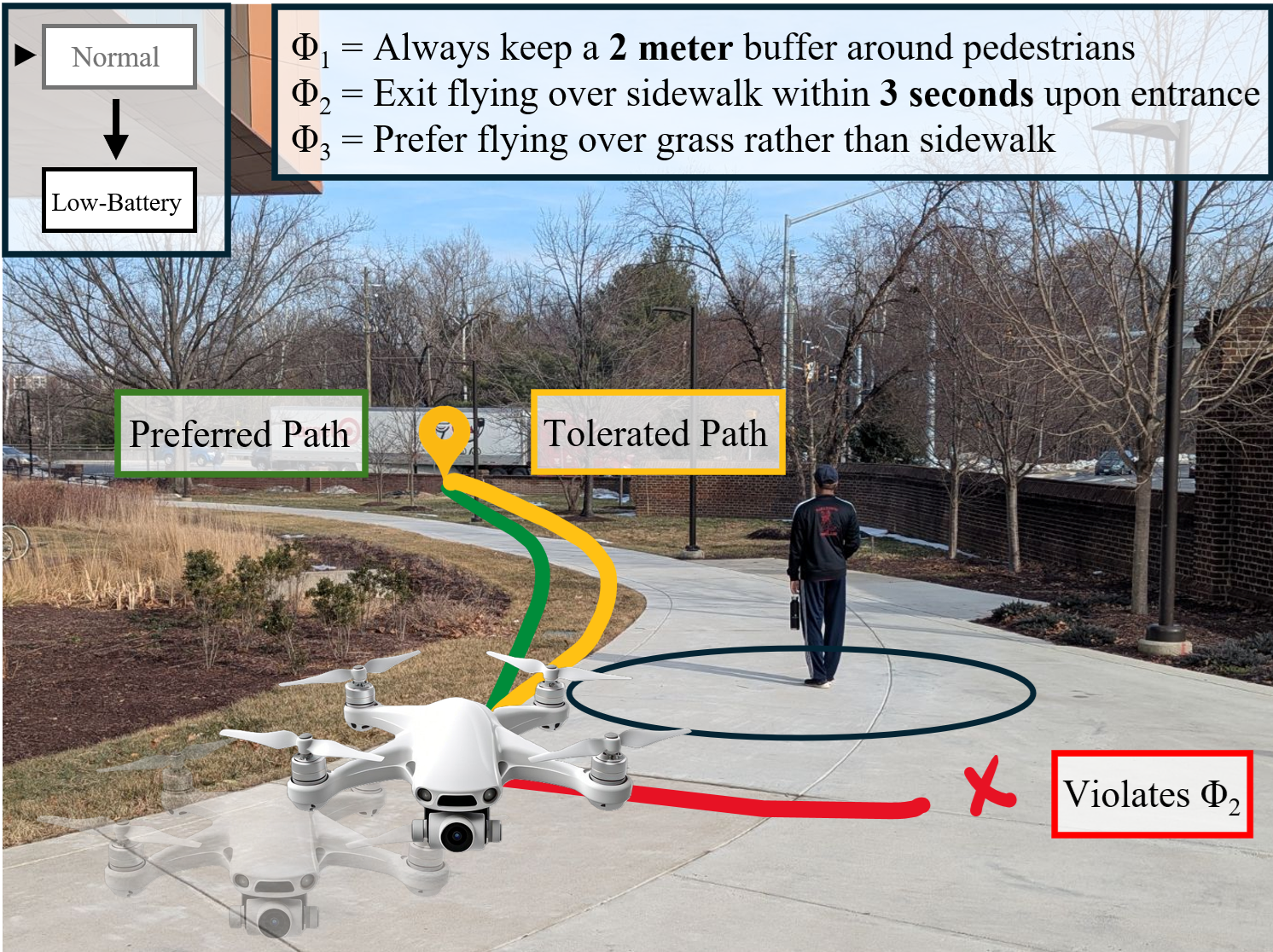}%
    \label{fig:bottom}
  }
  \caption{State-dependent navigation illustrates normal versus low-battery mode. The finite-state controller switches specifications $\phi_1 - \phi_3$, which changes which candidate paths are treated as preferred (green), valid and less-preferred/tolerated (yellow), or violating (red).}
  \label{fig:stacked}
\end{figure}

The architecture explicitly supports changes in robot operational states. We define two representative states: a \textbf{normal mode} with initial specifications and a degraded \textbf{low-battery mode} with stricter constraints on speed, obstacle clearance, and time-bounded behavior. Such mode changes illustrate how safety and performance requirements can adapt dynamically to external and internal system conditions. Table \ref{table:rules} lists the STL specifications for both modes and Figure \ref{fig:stacked} illustrates the state-change mechanism.

In this approach, we envision the transition to be manually triggered, though it could be automated through system state monitoring (e.g., battery triggers). When a state change occurs, inducing new specifications and desired robot behaviors, old specifications and cost maps are discarded. The robot pauses operation, recomputes its environmental cost map, and replans a path consistent with the new specifications. The state-switching mechanism demonstrates how tightening formal constraints affects the robot's reasoning process, cost structure, and resulting trajectory. 

\begin{table*}[h!]
\vspace{1.5mm}
\begin{tabular}{>{\raggedright\arraybackslash}p{3cm}|
                >{\raggedright\arraybackslash}p{6.8cm}|
                >{\raggedright\arraybackslash}p{6.8cm}}
\small
\setlength{\tabcolsep}{2pt}
\textbf{Rule Description}  & \textbf{Normal State}  & \textbf{Low-Battery State}\\
\hline
1. Global Speed Limit   & \makecell[l]{The speed must always be below 5 kph. \\ $G_{[0, \infty]} speed < 5 kph$}  & \makecell[l]{The speed must always be below 3 kph. \\ $G_{[0, \infty]} speed < 3 kph$}  \\[4ex]

2. Obstacle Avoidance with Buffer & \makecell[l]{Always keep 1 meter away from obstacles. \\ $G_{[0,\infty]} dist\_o \geq 1 m$} & \makecell[l]{Always keep 2 meters away from obstacles. \\ $G_{[0,\infty]} dist\_o \geq 2 m$ }  \\[4ex]

3. Sidewalk-Exit Temporal Rule & \makecell[l]{Exit flying over sidewalk within 5 seconds upon entrance. \\ $G_{[0,\infty]} status = sidewalk \rightarrow$ \\ $F_{[0,5]} status = \lnot sidewalk$ }  & \makecell[l]{Exit flying over sidewalk within 3 seconds upon entrance. \\ $G_{[0,\infty]} status = sidewalk \rightarrow$ \\$F_{[0,3]} status = \lnot sidewalk$ } \\[6ex]

4.1 Slow Before Stop & \makecell[l]{Slow to 5 kph within 5 seconds when nearing a stop sign, \\ then stop. \\ $ G_{[0,\infty]} stop\_obs \rightarrow F_{[0,5]} slow \land (F_{[0,5]} stop) $} & \textit{Same as the normal state.} \\[4ex]

4.2 Pause At Stop & \makecell[l]{The robot must stop at the sign for at least 3 seconds. \\ $G_{[0,\infty]} at\_stop \rightarrow G_{[0,3]} speed =0$  } & \textit{Same as the normal state.} \\[4ex]     

5. Preference to Avoid Pedestrian Sidewalk  & \makecell[l]{Prefer flying over grass rather than sidewalk. \\ \textit{Note: Soft preferences are encoded in the cost map.}} & \textit{Preference removed in the low-battery state.} \\ [4ex]
\end{tabular}

\caption{Table describing rules and soft operator-based preferences for both robot states.}
\label{table:rules}
\end{table*}

\subsection{Test \& Evaluation}
This architecture lends itself naturally to test and evaluation (T\&E) analysis because STL provides quantitative measures of specification satisfaction through robustness metrics and a cost map provides insights into navigation efficiency and risk. Each execution generates a final cost, trajectory data, system logs, and corresponding robustness values for each specification, thereby enabling rule monitoring of both frequency and severity of specification violations.

Beyond treating violations as a binary phenomenon, we monitor for violation metrics such as the average robustness margin to safety constraints, average violation magnitude, and worst-case robustness over a run, and aggregate this data across multiple trials. These aggregate statistics facilitate post-hoc analysis to tune rule preferences, reconstruct where and why violations occur, and a quantitative comparison between different system configurations.

Furthermore, by assigning priority levels to rules (e.g., ``hard" constraints like safety and speed limits versus ``soft" preferences such as path smoothness), we can evaluate trade-offs explicitly during operation. The system's robustness and adaptability to specification changes can then be analyzed both online and offline using the same family of robustness scores and rule-violation metrics.

Together, these metrics define an interpretable T\&E framework that clarifies how human language rules can be instantiated, monitored, and enforced within any perception-based robotic framework. This approach moves the navigation system toward a \textbf{correct-by-construction} design, providing a formal and empirically testable connection between human language rules, semantic understanding, and verified behavior of an autonomous robot.

%% file: Sections/4_Conclusion.tex
\section{Conclusion}

In this work, we proposed a theoretical framework that integrates human-provided safety rules into a Vision-and-Language Navigation stack for autonomous robots in novel environments. By grounding time-invariant Signal Temporal Logic (STL) specifications in VLM-based semantic segmentation, we obtain an environment cost map that encodes semantic preferences and safety margins. We then instantiate time-varying rules into a runtime assurance layer, which shapes candidate paths to satisfy a specified robustness threshold. In addition, we outline a theoretical T\&E methodology that supports both online and offline verification of navigation systems. 

Ongoing to this project is the instantiation of this framework into software for simulation-based experimentation. Future work aim to provide experimental hardware testing on a quadruped robot and UAV. We also plan to investigate automatically switching protocols from temporal logic systems for hybrid systems \cite{liuSynthesisReactiveSwitching2013}. Another direction is to evaluate VLM components for resiliency under varying environmental and sensing conditions as part of the offline, post-hoc analysis of state change mechanisms. Using this framework to evaluate current VLM-based navigation systems on their ability to follow language-based rules during planning is also a potential avenue of exploration. Finally, we note that this approach can be incorporated into classical navigation stacks, complementing existing infrastructure to provide higher-fidelity safety reasoning for autonomous robot navigation.